\pdfoutput=1

\documentclass[11pt]{article}

\usepackage[]{ACL2023}

\usepackage{times}
\usepackage{latexsym}
\usepackage{hyperref}
\usepackage{arydshln}
\usepackage{graphicx}

\usepackage[T1]{fontenc}

\usepackage[utf8]{inputenc}

\usepackage{microtype}

\usepackage{inconsolata}
\usepackage{multirow}
\usepackage{booktabs}

\title{Simple LLM Prompting is State-of-the-Art for Robust and Multilingual Dialogue Evaluation}

\author{John Mendonça\textsuperscript{1,2,\thanks{\hspace{0.2cm}Work conducted as a visiting scholar at CMU.}},  Patrícia Pereira\textsuperscript{1,2},\\ {\bf Helena Moniz\textsuperscript{1,3}}, {\bf João Paulo Carvalho\textsuperscript{1,2}}, {\bf Alon Lavie\textsuperscript{4,5}} \and {\bf Isabel Trancoso\textsuperscript{1,2}} \\
  \textsuperscript{1} INESC-ID, Lisbon \\
  \textsuperscript{2} Instituto Superior Técnico, University of Lisbon \\
  \textsuperscript{3} Faculdade de Letras, University of Lisbon \\
  \textsuperscript{4} Carnegie Mellon University, Pittsburgh \\
  \textsuperscript{5} Phrase, Pittsburgh \\
  \texttt{john.mendonca@inesc-id.pt} \\}

\begin{document}
\maketitle

\begin{abstract}

Despite significant research effort in the development of automatic dialogue evaluation metrics, little thought is given to evaluating dialogues other than in English. At the same time, ensuring metrics are invariant to semantically similar responses is also an overlooked topic. In order to achieve the desired properties of robustness and multilinguality for dialogue evaluation metrics, we propose a novel framework that takes advantage of the strengths of current evaluation models with the newly-established paradigm of prompting Large Language Models (LLMs). Empirical results show our framework achieves state of the art results in terms of mean Spearman correlation scores across several benchmarks and ranks \textbf{first place on both the Robust and Multilingual tasks} of the DSTC11 Track 4 “Automatic Evaluation Metrics for Open-Domain Dialogue Systems”, proving the evaluation capabilities of prompted LLMs.

\end{abstract}

\section{Introduction}

\begin{figure}[t]
  \centering
  \includegraphics[width=0.48\textwidth]{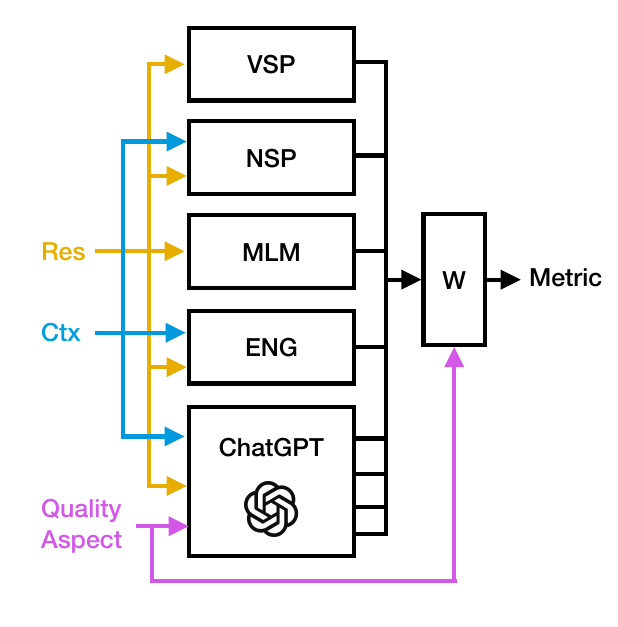}
  \caption{Proposed framework architecture. The \textbf{Response}, \textbf{Context} and \textbf{Quality Aspect} under evaluation are fed to the submetrics: \textbf{VSP} (Valid Sentence Prediction), \textbf{NSP} (Next Sentence Prediction), \textbf{MLM} (Masked Language Modelling), \textbf{ENG} (Engagement) and \textbf{ChatGPT}. Each submetric score is then weighted according to the aspect, yielding the final metric.}
  \label{fig:system}
\end{figure}

Automatic dialogue evaluation has largely been focused on evaluating select few languages. The main reason for this constraint is the lack of linguistic diversity in dialogue corpora, which leads to a lack of chatbots that cover other languages. As a result, the need for multilingual metrics has also been limited.
 
A possible solution to this issue is to leverage the latest batch of Large Language Models (LLMs) to synthetically generate multilingual dialogues. Some research has already been conducted to study the capabilities of these models \cite{guo2023close,bubeck2023sparks} and the consensus appears to be that these models have achieved a proxy of a \textit{formal} linguistic competence in the most studied languages. That is, its responses follow linguistic conventions and are fluent and grammatical, but they might be inaccurate or even hallucinate \citep{guerreiro2023hallucinations}. More importantly, pertaining to dialogue, they also show signs of \textit{functional} linguistic competence in its responses, i.e., discursive coherence, narrative structure and linguistic knowledge, even if not fully consistent (sometimes they do not consider context or situated information, and fail to adapt to users and domains). 

Irrespective of these models' limitations, it is clear their emergent capabilities allow for the development of chatbots with capabilities vastly beyond what earlier models were able to achieve. Yet, an interesting research question lingers: \textit{If these models are able to write responses that follow formal and functional linguistics rules, are they also capable of evaluating responses/dialogues in terms of these same rules?} Prior work has confirmed the language understanding capabilities of instruction-based LLMs for dialogue evaluation \cite{huynh2023understanding}. However, we are the first to study the evaluation capabilities of the newest batch of LLMs in terms of multilinguality and paraphrase robustness.

This paper presents our contribution to the DSTC11 track on Robust and Multilingual Automatic Evaluation Metrics for Open-Domain Dialogue Systems \cite{rodriguezcantelar2023robust}, where we participated in both the Multilingual and Robustness tasks. This track is an excellent venue to benchmark the capabilities of these new LLMs for dialogue evaluation, as it evaluates properties that have been observed in these models. We propose a comprehensive framework, incorporating earlier encoder-based approaches and \href{https://openai.com/blog/chatgpt}{ChatGPT}, as illustrated in Figure \ref{fig:system}. By combining multiple models and submetrics through ensembling, our approach aims to improve the performance and robustness of dialogue evaluation, ultimately contributing to the advancement of dialogue system research and development.

Overall, our contributions are the following:

\begin{itemize}
    \item We show that ChatGPT is a \textit{strong} evaluator of dialogues, outperforming typical encoder frameworks.
    \item We propose a new framework for dialogue evaluation that is multilingual and robust to paraphrases. In fact, our combined Encoder and ChatGPT framework ranks \textbf{1st place} on both the Multilingual and Robust metrics task.
    \item We discuss the outlook of Dialogue Evaluation in this new realm of LLMs.
    \item We open source the code and checkpoints of the submetrics at \url{github.com/johndmendonca/DialEvalML}.
\end{itemize}

\section{Related work}

\subsection{Automatic Evaluation Metrics}

Statistic-based metrics such as BLEU \citep{papineni-etal-2002-bleu}, ROUGE \cite{lin2004rouge}, and METEOR \citep{banerjee-lavie-2005-meteor}, are a popular choice to evaluate NLG (Natural Language Generation) models as they are easy to employ. These metrics assume valid responses have significant word-overlap with the ground truth. However, this is not a valid assumption for dialogue: there are many equally good responses for a single utterance. As such, the correlation with Human Evaluation (HE) annotations is very low for these metrics \citep{liu-etal-2016-evaluate}, and they cannot be used to evaluate models whenever a gold-response is not available.

Earlier learned metrics such as ADEM \cite{lowe2017towards} and RUBER \cite{tao2018ruber} explicitly predict HE annotations by initialising pretrained Recurrent Neural Network response generators. Unlike ADEM, which is trained with HE-annotated data in a supervised manner, RUBER leverages negative samples. In both cases, a reference response is used to score the candidate response. As such, these metrics still suffer the same issues as word-overlap metrics.

The primary motivation for the negative sampling approach in RUBER was the need for extensive HE annotations in ADEM. Approaches similar to this are now the norm for training open-domain dialogue evaluation metrics. By using well-defined self-supervised tasks which correlate well with their corresponding aspects, the annotation limitations are mostly circumvented. 

The most widely used self-supervised task is Next Sentence Prediction (NSP), as it is known to correlate well with HE that evaluate \textit{"Context Awareness"}. The typical approach is to finetune a pretrained encoder model with this automatically generated data \citep{mehri-eskenazi-2020-usr,phy-etal-2020-deconstruct,mendonca-etal-2022-qualityadapt,zhao,zhang2022mme}.
More complex approaches leverage graph representations to model dialogue interactions explicitly \cite{huang-etal-2020-grade, zhang-etal-2021-dynaeval}. Another typically employed self-supervised task is Valid Sentence Prediction (VSP), which uses word-level noising techniques to generate negative samples and correlates well with HE that evaluate \textit{Fluency} \citep{phy-etal-2020-deconstruct,mendonca-etal-2022-qualityadapt,zhang2022mme}.

Parallel to this trend, other annotation-free approaches in the literature have surfaced. For instance, qualities such as \textit{Specificity} correlate reasonably well with metrics obtained directly from the MLM (Masked Language Modelling) loss calculated using pretrained encoder models \citep{mehri-eskenazi-2020-usr,phy-etal-2020-deconstruct,zhang2022mme}. 

Given the multifaceted nature of dialogue, dialogue quality metrics typically employ a combination of submetrics. \citet{mehri-eskenazi-2020-unsupervised} leverage follow-up utterance from a pretrained decoder model to calculate 18 turn and dialogue-level submetrics, which are then used as inputs to a regression model for overall quality. In fact, Linear Regression is frequently used as a feature aggregation method in the literature \cite{jiang-etal-2022-im, mehri-eskenazi-2020-usr}. Alternatively, \citet{phy-etal-2020-deconstruct} propose a hierarchical composition where they incorporate the quality aspects together in a way that aspects in the lower hierarchy need to be satisfied before aspects higher up are considered. Also worth mentioning is the work of \citet{zhang2022mme}, which proposes the so called Correlation Re-scaling method. Here, the contribution of each aspect is calculated from the individual correlations of the submetrics, obtained from a subset of HE.

\subsection{Large Language Models}

The widespread use of LLMs was established, practically speaking, with the work of \citet{devlin-etal-2019-bert}, where a transformer architecture \cite{NIPS2017_3f5ee243} is pretrained with substantial amounts of unlabelled text with a Masked Language Modelling (MLM) objective. With this architecture, a new paradigm in NLP surfaced, where the adaptation to downstream tasks was conducted by finetuning the pretrained model with supervised data. Later on, GPT-3 \cite{brown2020language}, which is trained with an autoregressive objective, showed competitive results by leveraging few-shot prompting. Nevertheless, given their training objective function, it was difficult for autoregressive LLMs to successfully perform downstream NLP tasks without substantial prompt engineering. 

\citet{ouyang2022training} propose finetuning GPT-3 using a 3-step approach named Reinforcement Learning through Human Feedback (RLHF). In detail, the model is (1) initially finetuned using supervised data obtained from labelling prompts (SFT); (2) a reward model is trained using ranked responses given a prompt; (3) the policy is optimised against the reward model using the Proximal Policy Optimisation
reinforcement learning algorithm \citep{schulman2017proximal}. As a testament to the power of this approach, ChatGPT took the world by storm in late 2022 thanks to its incredible human-like generation capabilities. This was achieved by including dialogues in all steps of RLHF.

\section{Problem Formulation}

The main goal of this track was to develop and benchmark automatic open-ended dialogue evaluation metrics. Two tasks were proposed this year, Metrics for Multilingual Data and Robust metrics. For the Metrics for Multilingual Data task, participants were asked to construct quality metrics that perform well on a multilingual setup. For the the Robust metrics task, the goal was to develop metrics that perform robustly when evaluated over back-translated/paraphrased sentences in English.

In both tasks, the proposed metrics were evaluated at the turn and dialogue level, without access to a reference. In a \textit{turn-level} evaluation setting, the goal is, given prior dialogue history (frequently denoted as context) $c$ of varying amount of turns, and a response $r$, to learn a scoring function (also known as metric) that assigns a score $f(c,r) \rightarrow s$. Conversely, in a \textit{dialogue-level} evaluation setting, the goal is to evaluate the performance throughout the full dialogue. 

Irrespective of the level of evaluation, the proposed metrics' outputs are typically compared against HE annotations that use a Likert scale, where the lowest value means lowest quality and highest value maximum quality. For this track, the performance of these metrics was evaluated by calculating the Pearson correlation between the calculated score and HE.

\section{Methodology}

Our framework, which we call \textsc{DialEvalML}, can be viewed as a dual layered ensemble which are done at the \textbf{model} and \textbf{submetric} level, and that employ strong multilingual pretrained encoder and decoder models which were finetuned or prompted\footnote{We tried experimenting with metrics that use graph representations, but found implementing these metrics to be Multilingual and Robust, and including them in our framework, to be impractical, not to mention detrimental to performance in some instances.}. In this section, we describe the step-by-step process of \textsc{DialEvalML}, detailing the various components and methods employed.

\subsection{Submetrics}

Similar to other frameworks, including the best performing ones in last year's track \cite{zhang2022mme,jiang-etal-2022-im} which take inspiration from the works of \citet{phy-etal-2020-deconstruct,sinha-etal-2020-learning,mehri-eskenazi-2020-usr}, we employ several submetrics to evaluate dialogue responses -- ranging from zero-shot prediction using pretrained LLMs to trained models using self-supervised and supervised methods -- and weigh them according to the aspect we wish to predict. 

\subsubsection{VSP: Valid Sentence Prediction} 

Following \citet{sinha-etal-2020-learning}, we train a regression model that is optimised to differentiate between positive samples and synthetic negative samples. \textbf{Positive} samples are perturbed by randomly applying one of the following: (1) no perturbation, (2) punctuation removal, (3) stop-word removal. \textbf{Negative} samples are generated by randomly applying one of the following rules: (1) word reorder (shuffling the ordering of the words); (2) word-drop; and (3) word-repeat (randomly repeating words).

\subsubsection{NSP: Next Sentence Prediction} 

With the binary \textbf{NSP} (Next Sentence Prediction) task, the goal is to distinguish a positive example from a semantically negative one, given a context. We train a discriminative regression model using the following sampling strategy: \textbf{positive} responses are drawn directly from the dialog; \textbf{negative} responses are randomly selected and a token coverage test discards semantically similar sentences. All responses are processed using the positive-sample heuristic used by \textbf{VSP}.

For both tasks, the underlying goal is that paraphrased and/or translated responses should have the same coherence score as the original response, since they (in theory) convey the same message. In order to increase the robustness of our framework to paraphrased responses we propose a Siamese Neural Network. Simply put, we train an encoder model (denoted NSP-Siamese) to jointly optimise a Cosine Embedding Loss between the hidden states of the encoder model for the original and a paraphrase, and the individual errors between the predictions and the ground truth. We hypothesise this enables the model to compare the semantic coherence of the responses w.r.t the context, instead of more spurious features such as syntax. 

A similar approach could've been employed for multilingual metrics, however, scaling to more languages is computationally expensive: one would either need a new model for each language, or a training procedure requiring a forward pass for each language, for each example.

\subsubsection{MLM: Masked Language Modelling} 

Similar to \citet{mehri-eskenazi-2020-usr,phy-etal-2020-deconstruct}, we use a pretrained encoder model to calculate the MLM loss of all tokens of the response. The resulting \textbf{MLM} submetric is calculated as the sum of the individual losses.

\subsubsection{ENG: Engagement}

An important quality aspect of dialogue that is frequently overlooked is \textit{Engagement}. Some work attempt to equate this aspect with \textit{Specificity} and related metrics. However, we argue this is a reductive solution, as engagement is an abstract and multi-dimensional concept, thereby making a surface level evaluation of the response in terms of diversity insufficient.

As such, and following the methodology used for \textbf{VSP} and \textbf{NSP}, we train a discriminate model using RED (Reddit-based Engagement Dataset) \cite{xu-etal-2022-endex} which we then use as a submetric denoted in our framework as \textbf{ENG}. This dataset is sourced from Reddit and is curated using a novel distant-supervision framework. This framework aggregates emotional, attentional, behavioural and reply engagement onto a single score denoted \textsc{EnDex}, which then has a hyperparameter threshold applied to it to cluster posts into positive and negative samples.

\subsection{Exploiting Data Augmentation for Robust and Multilingual Evaluation}

The main novelty of this year's track is the release of training and development dialogue data that has been augmented with MT (Machine Translation) -- for the Multilingual task -- and Paraphrases -- for the Robust task. These augmentations are subsequently scored to determine similarity against the original data: for MT, several COMET QE (Quality Estimation) scores \cite{rei-etal-2020-unbabels,zerva-etal-2021-ist, rei-etal-2022-cometkiwi} were provided; for Paraphrases, the organisers provided cosine similarity scores of the sentence embeddings.

A naive approach to obtain competitive metrics in both tasks would be to simply introduce the full amount of augmented data during self-supervised and supervised training. However, \citet{mendoncaetal2023towards} showed that low quality augmentation affects the performance of models trained on MT augmented data, especially for \textbf{VSP}. Following this work, we select 5 and 75 \% of the best translated data (ranked using COMET QE) for training of the \textbf{VSP} and \textbf{NSP} models respectively. For \textbf{ENG}, we train different proportions of data and select the best performing ones.

\subsection{ChatGPT}

We briefly experimented with different prompts, and found the best performing prompt (irrespective of language) in a held-out internal set to be simply:

\begin{itemize}
    \item Turn-level: \textit{"Given the Context, evaluate from 1-5 the Response in terms of \{aspect\}. Provide a single score and nothing else."}\
    \item Dialogue-level: \textit{"Evaluate the following dialogue from 1-5 in terms of \{aspect\}. Provide a single score and nothing else."}
\end{itemize}

Unlike GPT-3, the API for ChatGPT does not output the log probabilities of the most likely tokens. As such, the measurement of quality is non-deterministic. We attempt to reduce output variability by reinforcing the desired output in the prompt (\textit{"Provide a single score and nothing else."}) and by setting the temperature to 0. We report a mean absolute deviation of 0.0182 across 3 runs when querying \textit{Appropriateness} on the provided \texttt{en/dailydialog-grade} dataset included in the development set. To facilitate ensembling in later stages, we normalise the predictions to [0,1].

The default processing step consists of searching for an integer in the response. However, there are some instances where ChatGPT fails to output the desired score: (1) When conducting dialogue level evaluation, the model sometimes outputs scores for each individual response. In these cases, we calculate the average score, similar to the dialogue-level encoder scores. (2) Less frequently, ChatGPT ignores the task and continues the conversation. Here, we prompt the model again until a score is provided.

\subsection{Submetric Ensembling}

Despite having a key role in NLG evaluation, HE has been performed while suffering from nontransparent and inconsistent annotation procedures. As such, annotations from different works one expects to report the same quality are frequently only nominal in nature. A good example is \textit{Coherence}, with some definitions referring to it as (1) semantic relevance with respect to a previous sentence; (2) a theme/topic; or even (3) \textit{Readability}, which is considered a different quality in other guidelines. \citet{howcroft-etal-2020-twenty} provides an in-depth survey of 165 NLG papers with human evaluations where these issues are highlighted.

Taking into account these facts, it is not clear we can successfully apply an empirical surjective mapping function from our submetrics to the quality aspects. Instead, we take a data-driven approach to generate this mapping, similar to the one proposed in \citet{zhang2022mme}. The main difference between the original Correlation Re-Scaling method and our approach is that, instead of zeroing the weights of submetrics that have a negative correlation with the given aspect, we take a probabilistic approach where we conduct a statistic significance test, i.e., we check if the $p$-value is higher than a given threshold. This ensures submetrics which are strongly and negatively correlated with the aspect (for example, \textbf{MLM} and \textit{Fluency}) are still included in the ensembling \footnote{For some annotations, none of the metrics were statistically significant. In these cases, we resort to the original proposed approach.}.

\subsection{Dialogue-level Evaluation}

We obtain dialogue-level quality predictions from the encoder models -- \textbf{NSP}, \textbf{VSP}, \textbf{MLM} and \textbf{ENG} -- by averaging the individual turn-level predictions. These are combined with the dialogue-level predictions obtained by prompting ChatGPT with the full dialogue in the prompt.

\section{Experiments}

\subsection{Datasets}

For data preprocessing we used \hyperlink{spacy.io}{spaCy}. For the \textbf{VSP} and \textbf{NSP} models, we followed prior work and base the self-supervised data on DailyDialog \cite{li-etal-2017-dailydialog}. For the language specific and multilingual models, we rank the translations using the provided WMT22 scores. Models using paraphrased responses are trained using the least similar responses (lowest score\footnote{We also trained models using the highest scoring responses and report lower performance. This is in line with our intuition that lower scoring responses are more diverse, and as such more informative for training.}).

The \textbf{ENG} model was trained using the RED dataset, more specifically on the 80k split with negative sampled data \cite{xu-etal-2022-endex}. Given it is an English dataset, we use \texttt{MBART50}\footnote{We chose \texttt{MBART50} as it is lightweight and open source.} \cite{liu-etal-2020-multilingual-denoising} to augment the original dataset with Spanish and Chinese MT. Finally, we score it using the \texttt{WMT20-COMET-QE-DA} model \cite{rei-etal-2020-unbabels}. For the paraphrase augmentation, we follow the organisers' approach of using Parrot Paraphraser \cite {prithivida2021parrot} and scoring the paraphrases with Cosine Similarity.

\subsection{Training and Hyperparameters}

We used \texttt{XLM-RoBERTa-large} \cite{conneau-etal-2020-unsupervised} as the encoder model for the experiments. This model is the multilingual version of \texttt{RoBERTa}, pretrained on CommonCrawl data containing 100 languages. We used a single Quadro RTX 6000 24GB GPU for the encoder experiments, and accessed ChatGPT (\texttt{gpt-3.5-turbo}) in late March using the OpenAI API.

For the \textbf{VSP}, \textbf{NSP} and \textbf{ENG} metrics, a token representing the speaker was added for each turn, and a maximum history length of 3 turns was used during training. For predictions in the development and test sets we include the full conversational context whenever possible. If it surpasses input size limitations, we iteratively remove turns from the context, starting from the oldest one. We applied a regression head consisting of a 2-layer MLP with a hidden size of 1024 and a hyperbolic tangent function as activation for prediction. All parameters were trained/finetuned using Adam optimiser \cite{DBLP:journals/corr/KingmaB14}. The fully finetuned models used a learning rate of 3e-6 and were trained for 3 epochs using a batch size of 16. Evaluation was conducted every 10,000 steps. The best performing model on the evaluation set was selected for testing. For the \textbf{MLM} metric, we used the existing LM head available in the Transformers library \cite{wolf-etal-2020-transformers}.

With respect to the model-level ensembling, we conduct simple unweighted averaging of the predictions of the models. For the submetric-level ensembling, we define the mask threshold as $p>0.05$ and square the correlations following \citet{zhang2022mme}. For testing, we define a mapping from the development quality aspects to the test-set aspects and obtain the final weights by averaging the weights obtained on the test set.

\begin{table}[h]
\centering
\scriptsize
\begin{tabular}{l|l|ccccc}
\toprule
\multicolumn{2}{l}{} &  \multicolumn{5}{c}{\textbf{Language}} \\\midrule
\textbf{Submetric} & \textbf{Model} & \textbf{EN} & \textbf{ES} & \textbf{ZH} & \textbf{PA} & \textbf{ALL} \\\midrule
\multirow{5}{*}{\textbf{VSP}} & \textbf{EN} & 0.195 & 0.173 & 0.161 & \textbf{0.067} & 0.149 \\
 & ES & 0.156 & 0.183 & 0.158 & 0.012 & 0.127 \\
 & ZH & 0.179 & 0.111 & 0.102 & 0.086 & 0.119 \\
 & \textbf{PA} & \textbf{0.212} & \textbf{0.193} & \textbf{0.198} & 0.062 & \textbf{0.166} \\
 & \textbf{ML5} & 0.195 & 0.168 & 0.157 & 0.040 & 0.140 \\\midrule
\multirow{5}{*}{\textbf{NSP}} & EN & 0.279 & 0.256 & 0.286 & 0.267 & 0.272 \\
 & ES & 0.266 & 0.257 & 0.282 & 0.251 & 0.264 \\
 & ZH & 0.246 & 0.238 & 0.298 & 0.232 & 0.254 \\
 & \textbf{PA} & \textbf{0.307} & 0.279 & 0.286 & \textbf{0.279} & 0.288 \\
 & \textbf{ML75} & 0.300 & \textbf{0.284} & \textbf{0.311} & 0.272 & \textbf{0.292} \\ \midrule
\multirow{6}{*}{\textbf{ENG}} & EN & 0.319 & 0.275 & 0.251 & 0.260 & 0.276 \\
 & ML5 & 0.310 & 0.268 & 0.214 & 0.275 & 0.267 \\
 & \textbf{ML10} & 0.334 & 0.296 & 0.243 & 0.279 & 0.288 \\
 & \textbf{ML20} & \textbf{0.379} & \textbf{0.324} & \textbf{0.274} & \textbf{0.316} & \textbf{0.324} \\
 & \textbf{ML50} & 0.340 & 0.263 & 0.258 & 0.289 & 0.287 \\
 & PA & 0.265 & 0.245 & 0.213 & 0.265 & 0.247 \\ \bottomrule
\end{tabular}
\caption{Spearman Correlation scores of our trained model variants on all Language benchmarks on the full development set. The best score for each submetric and language is highlighted in \textbf{bold}. Models included in the final ensemble are in \textbf{bold}, except for NSP, which also includes NSP-Siamese.}
\label{tab:ens}
\end{table}

\subsection{Model ensembling}

\begin{table*}[t]
\centering
\small
\begin{tabular}{lcccccccc}
\toprule
\textbf{Aspect} & \textbf{VSP}                  & \textbf{NSP}                  & \textbf{MLM}                  & \textbf{ENG}                   & \textbf{cGPT-A}                & \textbf{cGPT-R}                & \textbf{cGPT-C}                & \textbf{cGPT-G}                \\ \midrule
\textbf{Appropriateness}  & \cellcolor[HTML]{F1F8EE}0.039 & \cellcolor[HTML]{99C884}0.176 & \cellcolor[HTML]{FFFFFF}0.017 & \cellcolor[HTML]{EAF4E5}0.0511 & \cellcolor[HTML]{A0CC8D}0.165 & \cellcolor[HTML]{93C47D}\textbf{0.185} & \cellcolor[HTML]{96C681}0.181 & \cellcolor[HTML]{93C47D}\textbf{0.185} \\
\textbf{Relevance}   & \cellcolor[HTML]{FAFCF9}0.014 & \cellcolor[HTML]{93C47D}\textbf{0.214} & \cellcolor[HTML]{FFFFFF}0.003 & \cellcolor[HTML]{F5FAF3}0.023  & \cellcolor[HTML]{A1CC8E}0.188 & \cellcolor[HTML]{96C680}0.210  & \cellcolor[HTML]{AFD49F}0.160  & \cellcolor[HTML]{A0CB8C}0.190  \\
\textbf{Content Richness}  & \cellcolor[HTML]{B2D5A3}0.176 & \cellcolor[HTML]{E0EEDA}0.085 & \cellcolor[HTML]{B0D4A0}0.181 & \cellcolor[HTML]{93C47D}\textbf{0.238}  & \cellcolor[HTML]{F7FBF5}0.039 & \cellcolor[HTML]{FFFFFF}0.022 & \cellcolor[HTML]{A1CC8E}0.210  & \cellcolor[HTML]{F2F8F0}0.048 \\
\textbf{Grammatical Correctness}  & \cellcolor[HTML]{E4F0DE}0.021 & \cellcolor[HTML]{CFE5C5}0.084 & \cellcolor[HTML]{FFFFFF}-0.06 & \cellcolor[HTML]{D6E9CE}0.061  & \cellcolor[HTML]{9AC886}0.238 & \cellcolor[HTML]{99C784}0.242 & \cellcolor[HTML]{B6D8A8}0.155 & \cellcolor[HTML]{93C47D}\textbf{0.258} \\ \bottomrule

\end{tabular}
\caption{Calculated submetric weights of System 1 for test set quality aspects. Highest weight per aspect in \textbf{bold}.}
\label{tab:weights}
\end{table*}

In order to determine the best combination of models to include in our model ensemble, all encoder based models that require training were trained using different subsets of data. This includes the original (EN) English data, the corresponding augmentations in Chinese (ZH), Spanish (ES) and Paraphrases (PA) and the QE-ranked multilingual augmentation (MLXX) \footnote{We only include the best performing ML models.}. 

Spearman correlation results are presented in Table \ref{tab:ens}. For the \textbf{VSP} submetric, we note that the inclusion of translations is detrimental to performance. In fact, the best performing models are PA, followed by EN. This contrasts with \textbf{NSP}, where we observe that the inclusion of more translated data improves performance. For \textbf{ENG}, the best performance is obtained with 20\% of translated data. We include the 10 and 50\% models in our framework to take advantage of ensembling.

\begin{table*}[t]
\centering
\small
\begin{tabular}{l|cc|cc|cc|cc|cc}
 & \multicolumn{2}{c|}{\textbf{EN}} & \multicolumn{2}{c|}{\textbf{ZH}} & \multicolumn{2}{c|}{\textbf{ES}} & \multicolumn{2}{c|}{\textbf{ML-AVG}} & \multicolumn{2}{c}{\textbf{Rank}} \\ \hline
\textbf{Team} & \textbf{Turn} & \textbf{Dial} & \textbf{Turn} & \textbf{Dial} & \textbf{Turn} & \textbf{Dial} & \textbf{Turn} & \textbf{Dial} & \textbf{Turn} & \textbf{Dial} \\ \hline
Baseline (AM-FM) & 0.2940 & 0.2414 & 0.0753 & 0.4648 & 0.1826 & \textbf{0.8080} & 0.1840 & 0.5047 & 4 & 2 \\
Team 2 & 0.1469 & - & 0.1054 & - & 0.0808 & - & 0.1110 & - & 5 & - \\
Team 4 (us) & & & & & & & &  & 1 & 1\\
\hspace{0.2cm} \textit{- S1 (\textsc{DialEvalML})} & \textbf{\textit{0.4818}} & \textbf{\textit{0.5342}} & \textbf{\textit{0.3936}} & \textbf{\textit{0.7133}} & \textbf{\textit{0.5890}} & \textbf{\textit{0.8080}} & \textbf{\textit{0.4881}} & \textbf{\textit{0.6852}} &  &  \\
\hspace{0.2cm} - S2 & 0.2625 & 0.3295 & 0.3096 & 0.7030 & 0.5056 & 0.2500 & 0.3592 & 0.4275 &  & \\
\hspace{0.2cm} - S3 & 0.4795 & 0.5251 & 0.3656 & 0.6701 & 0.5409 & 0.8080 & 0.4620 & 0.6677 &  & \\
\hspace{0.2cm} - S4 & 0.4586 & 0.5039 & 0.3618 & 0.5859 & 0.5412 & 0.5915 & 0.4539 & 0.5604 &  & \\
Team 5 & 0.3702 & 0.1865 & 0.0701 & 0.1356 & 0.1983 & 0.6830 & 0.2129 & 0.3350 & 3 & 3 \\
Team 7 & 0.2214 & - & 0.3112 & - & 0.5644 & - & 0.3657 & - & 2 & - \\ \bottomrule
\end{tabular}
\caption{Average Spearman correlation across the 4 dimensions evaluated for the baseline Deep AM-FM \cite{am_fm} and all participating teams on the Task 1 (Multilingual metrics) test set. \textbf{Bold} denotes the best result for the corresponding Language, \textit{italic} denotes our best submission.}
\label{tab:task1_teams}
\end{table*}

\subsection{Track Results}
\label{sec:main_res}

For the track we submitted 4 different systems, exploring the contribution of the different components of our framework:

\begin{itemize}
    \item \textbf{System 1 (\textsc{DialEvalML})}: Submetric ensembling of ChatGPT + XLM-R. 
    \item \textbf{System 2}: Submetric ensembling of XLM-R.
    \item \textbf{System 3}: Submetric ensembling of ChatGPT.
    \item \textbf{System 4}: Direct mapping of ChatGPT submetrics.
\end{itemize}

Table \ref{tab:weights} identifies the turn-level weights calculated for testing for System 1.

\paragraph{Task 1: Multilingual Metrics}

The results for each team for Task 1 are presented in Table \ref{tab:task1_teams}, together with all of our submissions. In all languages at both the dialogue and turn level, our submissions vastly outperform others, with the exception of S2, which has comparable results with other participants. This clearly demonstrates the conversational understanding ChatGPT possesses. As expected, the best submission is S1, which conducts submetric ensembling with the XLM-R submetrics. This is followed by S3 and S4, which are exclusive ChatGPT submissions with and without ensembling, respectively.

\paragraph{Task 2: Robust Metrics}

The results for each team for Task 2 are presented in Table \ref{tab:task2_teams}. Similar to Task 1, in Task 2, our ChatGPT submissions outperform other teams. However, at the dialogue level, the best performing model is AM-FM.

\begin{table}[h]
\centering
\small
\begin{tabular}{l|cc}
\textbf{Team} & \textbf{Turn (rank)} & \textbf{Dial (rank)} \\ \hline
Baseline (AM-FM) & 0.3387 (4) & \textbf{0.4800} (1) \\
Team 1 & 0.1537 (6) & 0.1111 (4)\\
Team 3 & 0.2697 (5) & 0.2196 (3) \\
Team 4 (us) & &  \\
\hspace{0.2cm} \textit{- S1 (\textsc{DialEvalML})} & \textit{\textbf{0.4890} (1)} & \textit{0.3031 (2)} \\
\hspace{0.2cm} - S2 & \hspace{-0.45cm}0.3320  & \hspace{-0.4cm}0.2335 \\
\hspace{0.2cm} - S3 & \hspace{-0.45cm}0.4756  & \hspace{-0.4cm}0.2979  \\
\hspace{0.2cm} - S4 & \hspace{-0.45cm}0.4427 & \hspace{-0.4cm}0.2492  \\
Team 6 & 0.4190 (2) & - \\
Team 7 & 0.3833 (3) & - \\ \bottomrule
\end{tabular}
\caption{Average Spearman correlation and corresponding rank across the 4 dimensions evaluated for the baseline Deep AM-FM and all participating teams on the Task 2 (Robust metrics) test set. \textbf{Bold} denotes the best result for the corresponding Language, \textit{italic} denotes our best submission.}
\label{tab:task2_teams}
\end{table}

\subsection{Example predictions}
\label{sec:ex_preds}

\begin{table}[ht]
\centering
\small
\begin{tabular}{l}
\toprule

\textbf{Chatbot:} Tennis racquets used to be made from cloth\\ strips stuffed with feathers haha! Then eventually\\ evolved into today's technology\\
\textbf{Human:}	oh i never knew that\\
\textbf{Chatbot:} Indeed, there was also aluminium construction\\ improvements allowed for the introduction of oversized\\ rackets by 1975 which was manufactured by weed \\ \hdashline
\textbf{Label:} \hspace{0.6cm}\textbf{A:} 4.7	\hspace{0.3cm}\textbf{C:} 4.7	\hspace{0.3cm}\textbf{G:} 4.3	\hspace{0.3cm}\textbf{R:} 5.0 \\
\textbf{ChatGPT:} \hspace{0.1cm}\textbf{A:} 1.0	\hspace{0.3cm}\textbf{C:} 2.0	\hspace{0.3cm}\textbf{G:} 2.0	\hspace{0.3cm}\textbf{R:} 1.0 \\
\textit{"The mention of "weed" as the manufacturer may cause}\\ \textit{confusion or be perceived as inappropriate without}\\
\textit{further context."}\\
\textbf{Best:} \hspace{0.8cm}\textbf{A:} 2.2	\hspace{0.3cm}\textbf{C:} 2.6	\hspace{0.3cm}\textbf{G:} 1.9	\hspace{0.3cm}\textbf{R:} 2.2 \\ \midrule\midrule
\textbf{Human:}	Don't cry for breakfast.\\
\textbf{Chatbot:}	Cry t for comfort\\
\textbf{Human:}	In the end, young people are always\\ inexplicably sad in adolescence. Which one is Qibing?\\ \hdashline
\textbf{Label:} \hspace{0.6cm}\textbf{A:} -	\hspace{0.6cm}\textbf{C:} 5.0	\hspace{0.3cm}\textbf{G:} 5.0	\hspace{0.3cm}\textbf{R:} 3.0 \\
\textbf{ChatGPT:} \hspace{0.1cm}\textbf{A:} 1.0	\hspace{0.3cm}\textbf{C:} 1.0	\hspace{0.3cm}\textbf{G:} 1.0	\hspace{0.3cm}\textbf{R:} 1.0 \\
\textit{"The response does not directly relate to the context or}\\ 
\textit{provide a meaningful answer. It seems unrelated and out}\\ 
\textit{of place. The mention of "Qibing" without any}\\
\textit{explanation further adds to the confusion.}\\
\textbf{Best:} \hspace{0.8cm}\textbf{A:} 1.3	\hspace{0.3cm}\textbf{C:} 1.9	\hspace{0.3cm}\textbf{G:} 1.1	\hspace{0.3cm}\textbf{R:} 1.2 \\
\bottomrule

\end{tabular}
\caption{Example turn-level predictions for \textit{Appropriateness}, \textit{Content Richeness}, \textit{Grammatical Correctness} and \textit{Relevance}. We include the ChatGPT explanation for \textit{Appropriateness}.}
\label{tab:dialogue}
\end{table}

Given the widely publicised emergent capabilities of current LLMs, it is worthwhile exploring where their quality predictions diverge from the annotators. To do so, we checked all instances where ChatGPT (System 4) diverges from the Human Evaluation (HE) annotations by more than 3 points. In all of the detected examples, we noted ChatGPT consistently underestimated the quality of the response when compared to HE. 

We present in Table \ref{tab:dialogue} two representative examples. In the first example, we see that ChatGPT erroneously underestimates quality due to the inclusion of \textit{"weed"} in the response. We posit this is due to the RLHF finetuning, which conditions the model to avoid inappropriate or divisive topics. In the second example, we see ChatGPT has trouble understanding the conversation. Although one could argue the HE scores for \textit{Correctness} and \textit{Appropriateness} are too high, it seems clear the response is undeserving of a minimum score for all aspects. In fact, if one prompts the model to provide an explanation for \textit{Content Richness}, it replies the following: \textit{"The response attempts to provide some content related to the topic of adolescent sadness, but it is vague and lacks depth. The mention of "Qibing" without any explanation or context leaves the reader confused. The response could benefit from more specific and informative details about the topic to increase its content richness."}. However, if anything, the inclusion of the last sentence increases the richness of the response. Yet, it seems ChatGPT is conflating \textit{Content Richness} with \textit{Relevance}. We observe the same behaviour in all other instances we studied, and is in line with the submetric weights (Table \ref{tab:weights}).

\section{Discussions}
\label{sec:discussions}

The results from our work on both tasks (Section \ref{sec:main_res}) reveals that ChatGPT vastly outperforms typical encoder approaches that are trained to discriminate positive samples from artificially generated negative ones. It is important to note that, compared to the months worth of research dedicated to optimise our encoder models (including curation, training and selection), we were able to easily outperform all other teams and our own encoder models with a day's worth of prompt engineering. This is, in our opinion, a turning point in the paradigm of dialogue evaluation.

In any case, we do find instances where ChatGPT fails to accurately evaluate aspects of quality, as identified in Section \ref{sec:ex_preds}. Future research directions may attempt to tackle the issues of score calibration by providing prompts that include examples and/or explicitly provide guidelines for scoring.

However, given the current landscape on dialogue generation, and as our submission suggests, dialogue evaluation, it is important to reflect on the value of current quality estimation frameworks. One might argue performing HE or developing metrics that evaluate responses and/or dialogues in terms of linguistic competence (e.g. \textit{Grammatical Correctness} or \textit{Coherence}) is no longer informative for the current and future crop of LLMs. Besides becoming ever so clear that these models no longer output responses that are incoherent or incorrect, we are reaching the point where these models are better evaluators than humans themselves \cite{gilardi2023chatgpt}. As such, developing metrics that correlate well with HE is becoming increasingly questionable.

One of the main contention points w.r.t the deployment of these models to the public pertain to their "safety" and "trustworthiness". But while "trustworthiness" can be evaluated by connecting the outputs to external and verifiable sources, the notion of "safety" is much more ambiguous. \citet{kempt2023appropriateness} suggests considering Positionality, Acceptability, and Value Alignment (PAVA) as features chatbots should have to fulfil appropriateness requirements. However, automatically evaluating if a chatbot has these features using current dialogue evaluation protocols seems implausible. Instead, the development of challenge sets for validation (such as the ones proposed in \citealt{valmeekam2023planning}) appears to be the logical next step for evaluation of future chatbots\footnote{See \hyperlink{https://github.com/openai/evals}{OpenAI Evals} for recent collaborative research efforts in this direction.}.

\section{Conclusion}

This paper presents a novel open-domain and reference-free dialogue evaluation framework that leverages strong pretrained LLMs using finetuning and zero-shot prompting. These models, combined with effective ensembling strategies, substantially outperform the previous automatic evaluation paradigm of only training LMs with semisupervised training objectives. In fact, \textsc{DialEvalML} ranks 1st on both the Robust (1st turn-level, 2nd dialogue level) and Multilingual (1st on both levels) tasks of Track 4 at DSTC11. 

\section*{Acknowledgements}

This research was supported by the Portuguese Recovery and Resilience Plan through project C645008882-00000055 (Responsible.AI), and by national funds through \textit{Fundação para a Ciência e a Tecnologia} (FCT) with references PRT/BD/152198/2021 and UIDB/50021/2020, and by the P2020 program MAIA (LISBOA-01-0247-FEDER-045909).

\bibliography{anthology,custom}

\begin{thebibliography}{42}
\expandafter\ifx\csname natexlab\endcsname\relax\def\natexlab#1{#1}\fi

\bibitem[{Banerjee and Lavie(2005)}]{banerjee-lavie-2005-meteor}
Satanjeev Banerjee and Alon Lavie. 2005.
\newblock \href {https://aclanthology.org/W05-0909} {{METEOR}: An automatic
  metric for {MT} evaluation with improved correlation with human judgments}.
\newblock In \emph{Proceedings of the {ACL} Workshop on Intrinsic and Extrinsic
  Evaluation Measures for Machine Translation and/or Summarization}, pages
  65--72, Ann Arbor, Michigan. Association for Computational Linguistics.

\bibitem[{Brown et~al.(2020)Brown, Mann, Ryder, Subbiah, Kaplan, Dhariwal,
  Neelakantan, Shyam, Sastry, Askell et~al.}]{brown2020language}
Tom Brown, Benjamin Mann, Nick Ryder, Melanie Subbiah, Jared~D Kaplan, Prafulla
  Dhariwal, Arvind Neelakantan, Pranav Shyam, Girish Sastry, Amanda Askell,
  et~al. 2020.
\newblock Language models are few-shot learners.
\newblock \emph{Advances in neural information processing systems},
  33:1877--1901.

\bibitem[{Bubeck et~al.(2023)Bubeck, Chandrasekaran, Eldan, Gehrke, Horvitz,
  Kamar, Lee, Lee, Li, Lundberg, Nori, Palangi, Ribeiro, and
  Zhang}]{bubeck2023sparks}
Sébastien Bubeck, Varun Chandrasekaran, Ronen Eldan, Johannes Gehrke, Eric
  Horvitz, Ece Kamar, Peter Lee, Yin~Tat Lee, Yuanzhi Li, Scott Lundberg,
  Harsha Nori, Hamid Palangi, Marco~Tulio Ribeiro, and Yi~Zhang. 2023.
\newblock \href {http://arxiv.org/abs/2303.12712} {{Sparks of Artificial
  General Intelligence: Early experiments with GPT-4}}.

\bibitem[{Conneau et~al.(2020)Conneau, Khandelwal, Goyal, Chaudhary, Wenzek,
  Guzm{\'a}n, Grave, Ott, Zettlemoyer, and
  Stoyanov}]{conneau-etal-2020-unsupervised}
Alexis Conneau, Kartikay Khandelwal, Naman Goyal, Vishrav Chaudhary, Guillaume
  Wenzek, Francisco Guzm{\'a}n, Edouard Grave, Myle Ott, Luke Zettlemoyer, and
  Veselin Stoyanov. 2020.
\newblock \href {https://doi.org/10.18653/v1/2020.acl-main.747} {Unsupervised
  cross-lingual representation learning at scale}.
\newblock In \emph{Proceedings of the 58th Annual Meeting of the Association
  for Computational Linguistics}, pages 8440--8451, Online. Association for
  Computational Linguistics.

\bibitem[{Damodaran(2021)}]{prithivida2021parrot}
Prithiviraj Damodaran. 2021.
\newblock {Parrot: Paraphrase generation for NLU.}

\bibitem[{Devlin et~al.(2019)Devlin, Chang, Lee, and
  Toutanova}]{devlin-etal-2019-bert}
Jacob Devlin, Ming-Wei Chang, Kenton Lee, and Kristina Toutanova. 2019.
\newblock \href {https://doi.org/10.18653/v1/N19-1423} {{BERT}: Pre-training of
  deep bidirectional transformers for language understanding}.
\newblock In \emph{Proceedings of the 2019 Conference of the North {A}merican
  Chapter of the Association for Computational Linguistics: Human Language
  Technologies, Volume 1 (Long and Short Papers)}, pages 4171--4186,
  Minneapolis, Minnesota. Association for Computational Linguistics.

\bibitem[{Gilardi et~al.(2023)Gilardi, Alizadeh, and
  Kubli}]{gilardi2023chatgpt}
Fabrizio Gilardi, Meysam Alizadeh, and Maël Kubli. 2023.
\newblock \href {http://arxiv.org/abs/2303.15056} {{ChatGPT Outperforms
  Crowd-Workers for Text-Annotation Tasks}}.

\bibitem[{Guerreiro et~al.(2023)Guerreiro, Alves, Waldendorf, Haddow, Birch,
  Colombo, and Martins}]{guerreiro2023hallucinations}
Nuno~M. Guerreiro, Duarte Alves, Jonas Waldendorf, Barry Haddow, Alexandra
  Birch, Pierre Colombo, and André F.~T. Martins. 2023.
\newblock \href {http://arxiv.org/abs/2303.16104} {Hallucinations in large
  multilingual translation models}.

\bibitem[{Guo et~al.(2023)Guo, Zhang, Wang, Jiang, Nie, Ding, Yue, and
  Wu}]{guo2023close}
Biyang Guo, Xin Zhang, Ziyuan Wang, Minqi Jiang, Jinran Nie, Yuxuan Ding,
  Jianwei Yue, and Yupeng Wu. 2023.
\newblock \href {http://arxiv.org/abs/2301.07597} {{How Close is ChatGPT to
  Human Experts? Comparison Corpus, Evaluation, and Detection}}.

\bibitem[{Howcroft et~al.(2020)Howcroft, Belz, Clinciu, Gkatzia, Hasan,
  Mahamood, Mille, van Miltenburg, Santhanam, and
  Rieser}]{howcroft-etal-2020-twenty}
David~M. Howcroft, Anya Belz, Miruna-Adriana Clinciu, Dimitra Gkatzia, Sadid~A.
  Hasan, Saad Mahamood, Simon Mille, Emiel van Miltenburg, Sashank Santhanam,
  and Verena Rieser. 2020.
\newblock \href {https://aclanthology.org/2020.inlg-1.23} {Twenty years of
  confusion in human evaluation: {NLG} needs evaluation sheets and standardised
  definitions}.
\newblock In \emph{Proceedings of the 13th International Conference on Natural
  Language Generation}, pages 169--182, Dublin, Ireland. Association for
  Computational Linguistics.

\bibitem[{Huang et~al.(2020)Huang, Ye, Qin, Lin, and
  Liang}]{huang-etal-2020-grade}
Lishan Huang, Zheng Ye, Jinghui Qin, Liang Lin, and Xiaodan Liang. 2020.
\newblock \href {https://doi.org/10.18653/v1/2020.emnlp-main.742} {{GRADE}:
  Automatic graph-enhanced coherence metric for evaluating open-domain dialogue
  systems}.
\newblock In \emph{Proceedings of the 2020 Conference on Empirical Methods in
  Natural Language Processing (EMNLP)}, pages 9230--9240, Online. Association
  for Computational Linguistics.

\bibitem[{Huynh et~al.(2023)Huynh, Jiao, Gupta, Mehri, Bajaj, Chaudhary, and
  Eskenazi}]{huynh2023understanding}
Jessica Huynh, Cathy Jiao, Prakhar Gupta, Shikib Mehri, Payal Bajaj, Vishrav
  Chaudhary, and Maxine Eskenazi. 2023.
\newblock \href {http://arxiv.org/abs/2301.12004} {Understanding the
  effectiveness of very large language models on dialog evaluation}.

\bibitem[{Jiang et~al.(2022)Jiang, Ye, Rao, Wang, and
  Miao}]{jiang-etal-2022-im}
Zhihua Jiang, Guanghui Ye, Dongning Rao, Di~Wang, and Xin Miao. 2022.
\newblock \href {https://aclanthology.org/2022.emnlp-main.762}
  {{IM}{\textasciicircum}2: an interpretable and multi-category integrated
  metric framework for automatic dialogue evaluation}.
\newblock In \emph{Proceedings of the 2022 Conference on Empirical Methods in
  Natural Language Processing}, pages 11091--11103, Abu Dhabi, United Arab
  Emirates. Association for Computational Linguistics.

\bibitem[{Kempt et~al.(2023)Kempt, Lavie, and Nagel}]{kempt2023appropriateness}
Hendrik Kempt, Alon Lavie, and Saskia~K. Nagel. 2023.
\newblock \href {http://arxiv.org/abs/2304.14553} {Appropriateness is all you
  need!}

\bibitem[{Kingma and Ba(2015)}]{DBLP:journals/corr/KingmaB14}
Diederik~P. Kingma and Jimmy Ba. 2015.
\newblock \href {http://arxiv.org/abs/1412.6980} {Adam: {A} method for
  stochastic optimization}.
\newblock In \emph{3rd International Conference on Learning Representations,
  {ICLR} 2015, San Diego, CA, USA, May 7-9, 2015, Conference Track
  Proceedings}.

\bibitem[{Li et~al.(2017)Li, Su, Shen, Li, Cao, and
  Niu}]{li-etal-2017-dailydialog}
Yanran Li, Hui Su, Xiaoyu Shen, Wenjie Li, Ziqiang Cao, and Shuzi Niu. 2017.
\newblock \href {https://aclanthology.org/I17-1099} {{D}aily{D}ialog: A
  manually labelled multi-turn dialogue dataset}.
\newblock In \emph{Proceedings of the Eighth International Joint Conference on
  Natural Language Processing (Volume 1: Long Papers)}, pages 986--995, Taipei,
  Taiwan. Asian Federation of Natural Language Processing.

\bibitem[{Lin(2004)}]{lin2004rouge}
Chin-Yew Lin. 2004.
\newblock Rouge: A package for automatic evaluation of summaries.
\newblock In \emph{Text summarization branches out}, pages 74--81.

\bibitem[{Liu et~al.(2016)Liu, Lowe, Serban, Noseworthy, Charlin, and
  Pineau}]{liu-etal-2016-evaluate}
Chia-Wei Liu, Ryan Lowe, Iulian Serban, Mike Noseworthy, Laurent Charlin, and
  Joelle Pineau. 2016.
\newblock \href {https://doi.org/10.18653/v1/D16-1230} {How {NOT} to evaluate
  your dialogue system: An empirical study of unsupervised evaluation metrics
  for dialogue response generation}.
\newblock In \emph{Proceedings of the 2016 Conference on Empirical Methods in
  Natural Language Processing}, pages 2122--2132, Austin, Texas. Association
  for Computational Linguistics.

\bibitem[{Liu et~al.(2020)Liu, Gu, Goyal, Li, Edunov, Ghazvininejad, Lewis, and
  Zettlemoyer}]{liu-etal-2020-multilingual-denoising}
Yinhan Liu, Jiatao Gu, Naman Goyal, Xian Li, Sergey Edunov, Marjan
  Ghazvininejad, Mike Lewis, and Luke Zettlemoyer. 2020.
\newblock \href {https://doi.org/10.1162/tacl_a_00343} {Multilingual denoising
  pre-training for neural machine translation}.
\newblock \emph{Transactions of the Association for Computational Linguistics},
  8:726--742.

\bibitem[{Lowe et~al.(2017)Lowe, Noseworthy, Serban, Angelard-Gontier, Bengio,
  and Pineau}]{lowe2017towards}
Ryan Lowe, Michael Noseworthy, Iulian~Vlad Serban, Nicolas Angelard-Gontier,
  Yoshua Bengio, and Joelle Pineau. 2017.
\newblock Towards an automatic turing test: Learning to evaluate dialogue
  responses.
\newblock In \emph{Proceedings of the 55th Annual Meeting of the Association
  for Computational Linguistics (Volume 1: Long Papers)}, pages 1116--1126.

\bibitem[{Mehri and
  Eskenazi(2020{\natexlab{a}})}]{mehri-eskenazi-2020-unsupervised}
Shikib Mehri and Maxine Eskenazi. 2020{\natexlab{a}}.
\newblock \href {https://aclanthology.org/2020.sigdial-1.28} {Unsupervised
  evaluation of interactive dialog with {D}ialo{GPT}}.
\newblock In \emph{Proceedings of the 21th Annual Meeting of the Special
  Interest Group on Discourse and Dialogue}, pages 225--235, 1st virtual
  meeting. Association for Computational Linguistics.

\bibitem[{Mehri and Eskenazi(2020{\natexlab{b}})}]{mehri-eskenazi-2020-usr}
Shikib Mehri and Maxine Eskenazi. 2020{\natexlab{b}}.
\newblock \href {https://doi.org/10.18653/v1/2020.acl-main.64} {{USR}: An
  unsupervised and reference free evaluation metric for dialog generation}.
\newblock In \emph{Proceedings of the 58th Annual Meeting of the Association
  for Computational Linguistics}, pages 681--707, Online. Association for
  Computational Linguistics.

\bibitem[{Mendonca et~al.(2022)Mendonca, Lavie, and
  Trancoso}]{mendonca-etal-2022-qualityadapt}
John Mendonca, Alon Lavie, and Isabel Trancoso. 2022.
\newblock \href {https://aclanthology.org/2022.sigdial-1.9} {{Q}uality{A}dapt:
  an automatic dialogue quality estimation framework}.
\newblock In \emph{Proceedings of the 23rd Annual Meeting of the Special
  Interest Group on Discourse and Dialogue}, pages 83--90, Edinburgh, UK.
  Association for Computational Linguistics.

\bibitem[{Mendonca et~al.(2023)Mendonca, Lavie, and
  Trancoso}]{mendoncaetal2023towards}
John Mendonca, Alon Lavie, and Isabel Trancoso. 2023.
\newblock Towards multilingual automatic open-domain dialogue evaluation.
\newblock In \emph{Proceedings of the 24th Annual Meeting of the Special
  Interest Group on Discourse and Dialogue}, Prague, Czechia. Association for
  Computational Linguistics.

\bibitem[{Ouyang et~al.(2022)Ouyang, Wu, Jiang, Almeida, Wainwright, Mishkin,
  Zhang, Agarwal, Slama, Ray, Schulman, Hilton, Kelton, Miller, Simens, Askell,
  Welinder, Christiano, Leike, and Lowe}]{ouyang2022training}
Long Ouyang, Jeff Wu, Xu~Jiang, Diogo Almeida, Carroll~L. Wainwright, Pamela
  Mishkin, Chong Zhang, Sandhini Agarwal, Katarina Slama, Alex Ray, John
  Schulman, Jacob Hilton, Fraser Kelton, Luke Miller, Maddie Simens, Amanda
  Askell, Peter Welinder, Paul Christiano, Jan Leike, and Ryan Lowe. 2022.
\newblock \href {http://arxiv.org/abs/2203.02155} {Training language models to
  follow instructions with human feedback}.

\bibitem[{Papineni et~al.(2002)Papineni, Roukos, Ward, and
  Zhu}]{papineni-etal-2002-bleu}
Kishore Papineni, Salim Roukos, Todd Ward, and Wei-Jing Zhu. 2002.
\newblock \href {https://doi.org/10.3115/1073083.1073135} {{B}leu: a method for
  automatic evaluation of machine translation}.
\newblock In \emph{Proceedings of the 40th Annual Meeting of the Association
  for Computational Linguistics}, pages 311--318, Philadelphia, Pennsylvania,
  USA. Association for Computational Linguistics.

\bibitem[{Phy et~al.(2020)Phy, Zhao, and Aizawa}]{phy-etal-2020-deconstruct}
Vitou Phy, Yang Zhao, and Akiko Aizawa. 2020.
\newblock \href {https://doi.org/10.18653/v1/2020.coling-main.368} {Deconstruct
  to reconstruct a configurable evaluation metric for open-domain dialogue
  systems}.
\newblock In \emph{Proceedings of the 28th International Conference on
  Computational Linguistics}, pages 4164--4178, Barcelona, Spain (Online).
  International Committee on Computational Linguistics.

\bibitem[{Rei et~al.(2020)Rei, Stewart, Farinha, and
  Lavie}]{rei-etal-2020-unbabels}
Ricardo Rei, Craig Stewart, Ana~C Farinha, and Alon Lavie. 2020.
\newblock \href {https://aclanthology.org/2020.wmt-1.101} {Unbabel{'}s
  participation in the {WMT}20 metrics shared task}.
\newblock In \emph{Proceedings of the Fifth Conference on Machine Translation},
  pages 911--920, Online. Association for Computational Linguistics.

\bibitem[{Rei et~al.(2022)Rei, Treviso, Guerreiro, Zerva, Farinha, Maroti,
  C.~de Souza, Glushkova, Alves, Coheur, Lavie, and
  Martins}]{rei-etal-2022-cometkiwi}
Ricardo Rei, Marcos Treviso, Nuno~M. Guerreiro, Chrysoula Zerva, Ana~C Farinha,
  Christine Maroti, Jos{\'e}~G. C.~de Souza, Taisiya Glushkova, Duarte Alves,
  Luisa Coheur, Alon Lavie, and Andr{\'e} F.~T. Martins. 2022.
\newblock \href {https://aclanthology.org/2022.wmt-1.60} {{C}omet{K}iwi:
  {IST}-unbabel 2022 submission for the quality estimation shared task}.
\newblock In \emph{Proceedings of the Seventh Conference on Machine Translation
  (WMT)}, pages 634--645, Abu Dhabi, United Arab Emirates (Hybrid). Association
  for Computational Linguistics.

\bibitem[{Rodríguez-Cantelar et~al.(2023)Rodríguez-Cantelar, Zhang, Tang,
  Shi, Ghazarian, Sedoc, D'Haro, and Rudnicky}]{rodriguezcantelar2023robust}
Mario Rodríguez-Cantelar, Chen Zhang, Chengguang Tang, Ke~Shi, Sarik
  Ghazarian, João Sedoc, Luis~Fernando D'Haro, and Alexander Rudnicky. 2023.
\newblock Overview of robust and multilingual automatic evaluation metrics for
  open-domain dialogue systems at dstc 11 track 4.
\newblock In \emph{DSTC11: The Eleventh Dialog System Technology Challenge},
  24th Meeting of the Special Interest Group on Discourse and Dialogue
  (SIGDIAL), Prague, Czechia.

\bibitem[{Schulman et~al.(2017)Schulman, Wolski, Dhariwal, Radford, and
  Klimov}]{schulman2017proximal}
John Schulman, Filip Wolski, Prafulla Dhariwal, Alec Radford, and Oleg Klimov.
  2017.
\newblock \href {http://arxiv.org/abs/1707.06347} {{Proximal Policy
  Optimization Algorithms}}.

\bibitem[{Sinha et~al.(2020)Sinha, Parthasarathi, Wang, Lowe, Hamilton, and
  Pineau}]{sinha-etal-2020-learning}
Koustuv Sinha, Prasanna Parthasarathi, Jasmine Wang, Ryan Lowe, William~L.
  Hamilton, and Joelle Pineau. 2020.
\newblock \href {https://doi.org/10.18653/v1/2020.acl-main.220} {Learning an
  unreferenced metric for online dialogue evaluation}.
\newblock In \emph{Proceedings of the 58th Annual Meeting of the Association
  for Computational Linguistics}, pages 2430--2441, Online. Association for
  Computational Linguistics.

\bibitem[{Tao et~al.(2018)Tao, Mou, Zhao, and Yan}]{tao2018ruber}
Chongyang Tao, Lili Mou, Dongyan Zhao, and Rui Yan. 2018.
\newblock Ruber: An unsupervised method for automatic evaluation of open-domain
  dialog systems.
\newblock In \emph{Proceedings of the AAAI Conference on Artificial
  Intelligence}, volume~32.

\bibitem[{Valmeekam et~al.(2023)Valmeekam, Sreedharan, Marquez, Olmo, and
  Kambhampati}]{valmeekam2023planning}
Karthik Valmeekam, Sarath Sreedharan, Matthew Marquez, Alberto Olmo, and
  Subbarao Kambhampati. 2023.
\newblock \href {http://arxiv.org/abs/2302.06706} {{On the Planning Abilities
  of Large Language Models (A Critical Investigation with a Proposed
  Benchmark)}}.

\bibitem[{Vaswani et~al.(2017)Vaswani, Shazeer, Parmar, Uszkoreit, Jones,
  Gomez, Kaiser, and Polosukhin}]{NIPS2017_3f5ee243}
Ashish Vaswani, Noam Shazeer, Niki Parmar, Jakob Uszkoreit, Llion Jones,
  Aidan~N Gomez, \L~ukasz Kaiser, and Illia Polosukhin. 2017.
\newblock \href
  {https://proceedings.neurips.cc/paper_files/paper/2017/file/3f5ee243547dee91fbd053c1c4a845aa-Paper.pdf}
  {Attention is all you need}.
\newblock In \emph{Advances in Neural Information Processing Systems},
  volume~30. Curran Associates, Inc.

\bibitem[{Wolf et~al.(2020)Wolf, Debut, Sanh, Chaumond, Delangue, Moi, Cistac,
  Rault, Louf, Funtowicz, Davison, Shleifer, von Platen, Ma, Jernite, Plu, Xu,
  Le~Scao, Gugger, Drame, Lhoest, and Rush}]{wolf-etal-2020-transformers}
Thomas Wolf, Lysandre Debut, Victor Sanh, Julien Chaumond, Clement Delangue,
  Anthony Moi, Pierric Cistac, Tim Rault, Remi Louf, Morgan Funtowicz, Joe
  Davison, Sam Shleifer, Patrick von Platen, Clara Ma, Yacine Jernite, Julien
  Plu, Canwen Xu, Teven Le~Scao, Sylvain Gugger, Mariama Drame, Quentin Lhoest,
  and Alexander Rush. 2020.
\newblock \href {https://doi.org/10.18653/v1/2020.emnlp-demos.6} {Transformers:
  State-of-the-art natural language processing}.
\newblock In \emph{Proceedings of the 2020 Conference on Empirical Methods in
  Natural Language Processing: System Demonstrations}, pages 38--45, Online.
  Association for Computational Linguistics.

\bibitem[{Xu et~al.(2022)Xu, Liu, Harel-Canada, Chandra, and
  Peng}]{xu-etal-2022-endex}
Guangxuan Xu, Ruibo Liu, Fabrice Harel-Canada, Nischal~Reddy Chandra, and
  Nanyun Peng. 2022.
\newblock \href {https://aclanthology.org/2022.findings-emnlp.359} {{E}n{D}ex:
  Evaluation of dialogue engagingness at scale}.
\newblock In \emph{Findings of the Association for Computational Linguistics:
  EMNLP 2022}, pages 4884--4893, Abu Dhabi, United Arab Emirates. Association
  for Computational Linguistics.

\bibitem[{Zerva et~al.(2021)Zerva, van Stigt, Rei, Farinha, Ramos, C.~de Souza,
  Glushkova, Vera, Kepler, and Martins}]{zerva-etal-2021-ist}
Chrysoula Zerva, Daan van Stigt, Ricardo Rei, Ana~C Farinha, Pedro Ramos,
  Jos{\'e}~G. C.~de Souza, Taisiya Glushkova, Miguel Vera, Fabio Kepler, and
  Andr{\'e} F.~T. Martins. 2021.
\newblock \href {https://aclanthology.org/2021.wmt-1.102} {{IST}-unbabel 2021
  submission for the quality estimation shared task}.
\newblock In \emph{Proceedings of the Sixth Conference on Machine Translation},
  pages 961--972, Online. Association for Computational Linguistics.

\bibitem[{Zhang et~al.(2021{\natexlab{a}})Zhang, Chen, D{'}Haro, Zhang,
  Friedrichs, Lee, and Li}]{zhang-etal-2021-dynaeval}
Chen Zhang, Yiming Chen, Luis~Fernando D{'}Haro, Yan Zhang, Thomas Friedrichs,
  Grandee Lee, and Haizhou Li. 2021{\natexlab{a}}.
\newblock \href {https://doi.org/10.18653/v1/2021.acl-long.441} {{D}yna{E}val:
  Unifying turn and dialogue level evaluation}.
\newblock In \emph{Proceedings of the 59th Annual Meeting of the Association
  for Computational Linguistics and the 11th International Joint Conference on
  Natural Language Processing (Volume 1: Long Papers)}, pages 5676--5689,
  Online. Association for Computational Linguistics.

\bibitem[{Zhang et~al.(2021{\natexlab{b}})Zhang, D'Haro, Banchs, Friedrichs,
  and Li}]{am_fm}
Chen Zhang, Luis~Fernando D'Haro, Rafael~E. Banchs, Thomas Friedrichs, and
  Haizhou Li. 2021{\natexlab{b}}.
\newblock \href {https://doi.org/10.1007/978-981-15-8395-7_5} {\emph{Deep
  AM-FM: Toolkit for Automatic Dialogue Evaluation}}, pages 53--69. Springer
  Singapore, Singapore.

\bibitem[{Zhang et~al.(2022)Zhang, Hu, Yu, Wang, Han, Liu, and
  Yuan}]{zhang2022mme}
Pengfei Zhang, Xiaohui Hu, Kaidong Yu, Jian Wang, Song Han, Cao Liu, and
  Chunyang Yuan. 2022.
\newblock {MME-CRS: Multi-Metric Evaluation Based on Correlation Re-Scaling for
  Evaluating Open-Domain Dialogue}.
\newblock \emph{arXiv preprint arXiv:2206.09403}.

\bibitem[{Zhao et~al.(2020)Zhao, Lala, and Kawahara}]{zhao}
Tianyu Zhao, Divesh Lala, and Tatsuya Kawahara. 2020.
\newblock Designing precise and robust dialogue response evaluators.
\newblock In \emph{Proceedings of the 58th Annual Meeting of the Association
  for Computational Linguistics}, pages 26--33, Online. Association for
  Computational Linguistics.

\end{thebibliography}
\bibliographystyle{acl_natbib}

\end{document}